%% file: main.tex
\documentclass{article} 
\usepackage{colm2024_conference}
\usepackage[T1]{tipa}
\usepackage{microtype}
\usepackage{hyperref}
\usepackage{url}
\usepackage{caption}
\usepackage{wrapfig}
\usepackage{booktabs}
\usepackage{graphicx}
\usepackage{tcolorbox}
\definecolor{darkblue}{rgb}{0, 0, 0.5}
\hypersetup{colorlinks=true, citecolor=darkblue, linkcolor=darkblue, urlcolor=darkblue}
\definecolor{junebud}{rgb}{0.74, 0.85, 0.34}
\definecolor{flamingopink}{rgb}{0.99, 0.56, 0.67}

\newcommand{\name}[1]{PhonologyBench}

\title{\name{}: Evaluating Phonological Skills of Large Language Models}

\colmfinalcopy
\author{Ashima Suvarna, Harshita Khandelwal, Nanyun Peng \\
Department of Computer Science\\
University of California, Los Angeles\\
\texttt{\{asuvarna31,nanyunpeng\}@cs.ucla.edu} \\
\texttt{\{harshitaskh\}@g.ucla.edu} \\
}

\definecolor{myboxcolor}{RGB}{245,245,245} 
\definecolor{myframe}{RGB}{0,0,128} 

\newtcolorbox{mybody}{
  colback=myboxcolor,
  colframe=myframe,
  halign=left,
  boxrule=1pt, 
  left=1pt,
  right=1pt,
  top=1pt,
  bottom=1pt,
}
\usepackage{xspace}
\newcommand{\llama}{LLaMA-2-13B-Chat\xspace}

\begin{document}

\maketitle

\input{sections/abstract}
\input{sections/introduction}
\input{sections/related_work}
\input{sections/benchmark}

\input{sections/experimental_results}

\input{sections/conclusion}

\bibliography{custom}
\bibliographystyle{colm2024_conference}

\appendix
\input{sections/appendix}

\end{document}

%% file: sections/abstract.tex
\begin{abstract}
Phonology, the study of speech's structure and pronunciation rules, is a critical yet often overlooked component in Large Language Model (LLM) research. LLMs are widely used in various downstream applications that leverage phonology such as educational tools and poetry generation. Moreover, LLMs can potentially learn imperfect associations between orthographic and phonological forms from the training data. Thus, it is imperative to benchmark the phonological skills of LLMs. To this end, we present \name{}, a novel benchmark consisting of three diagnostic tasks designed to explicitly test the phonological skills of LLMs in English: grapheme-to-phoneme conversion, syllable counting, and rhyme word generation. Despite having no access to speech data, LLMs showcased notable performance on the \name{} tasks. However, we observe a significant gap of 17\% and 45\% on Rhyme Word Generation and Syllable counting, respectively, when compared to humans. Our findings underscore the importance of studying LLM performance on phonological tasks that inadvertently impact real-world applications. Furthermore, we encourage researchers to choose LLMs that perform well on the phonological task that is closely related to the downstream application since we find that no single model consistently outperforms the others on all the tasks.

\end{abstract}

%% file: sections/introduction.tex
\section{Introduction}

Large language models (LLMs) are a class of deep generative models \citep{ruthotto2021introduction,bond2021deep} that are trained on an internet-scale text corpora (e.g., two trillion tokens).  To assess their usefulness in the real-world scenarios, prior work has focused on holistic evaluation of LLMs \citep{liang2022holistic}. For instance, they are evaluated on tasks that require syntactic and semantic understanding of the language such as summarization, rationale generation, story generation and question answering ~\citep{liang2022holistic,zheng2023judging, valmeekam2023large, bang2023multitask, open-llm-leaderboard, qin2023chatgpt, Koco__2023, sun2023evaluating}. However, there are various text-based tasks of practical importance that require joint understanding of the written and spoken language such as poetry generation ~\citep{ormazabal2022poelm,henderson1965haiku, suzuki1985role}. In particular, these tasks require the model to have phonological skills i.e., understanding the patterns of speech units and rules governing pronunciation in language. While LLMs have been adopted to perform text-based tasks that require phonological skills ~\citep{ding2024songcomposer, kwon2023interfaces}, it is unclear to what extent they acquire phonological skills by training on large-scale text data, without access to speech data.

\begin{figure}[t]
\begin{center}
\includegraphics[scale=0.3]{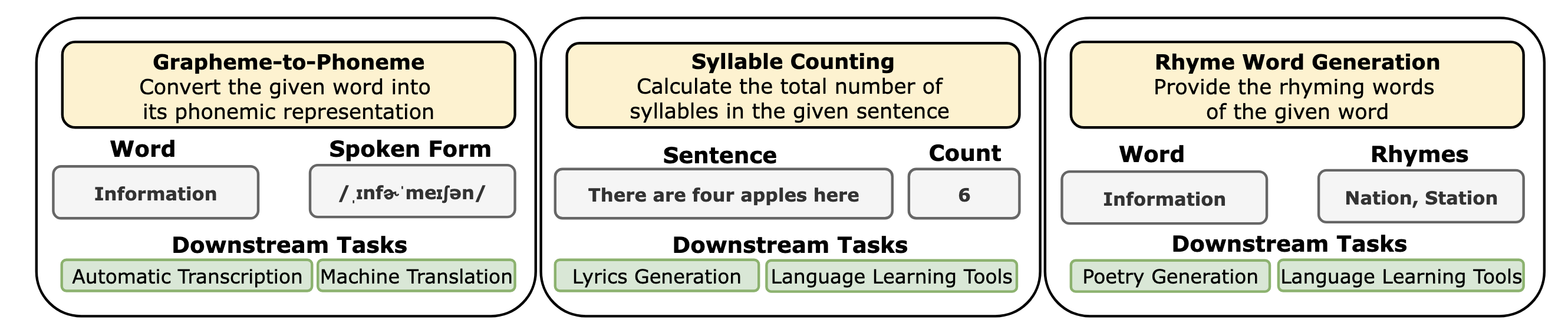}
\end{center}
\caption{Overview of tasks comprising \name{}. a) Grapheme-to-Phoneme conversion translates written language into phonetic script b) Syllable Counting enumerates the syllables in a sentence c) Rhyme Word Generation identifies rhyming counterparts for a given word. Each task is presented with an example and various downstream tasks.}
\label{fig:main}
\end{figure}

Despite being trained solely on textual data, LLMs have been applied to many tasks that benefit from a deep understanding of phonology. These applications include poetry generation, song writing, machine translation, and language learning ~\citep{ding2024songcomposer, kwon2023interfaces, yu2024token}. For example, the generation of poetry and lyrics leverages the models' knowledge of rhyme and meter to produce rhythmically engaging content. Similarly, phonetic transcriptions from related Dravidian languages can enhance the accuracy of multilingual machine translations~\citep{chakravarthi2019multilingual}. We argue that due to their extensive training and further alignment, LLMs may leverage the learnt associations between written and spoken forms to accomplish these tasks. However, due to the lack of deep phonological understanding, they fail to perform phonological reasoning tasks in open-world scenarios ~\citep{peng2023spoken}. For example, although LLMs are popular for composing poetry, machine-generated poetry lacks diverse rhyming schemes and is highly criticised by experts ~\citep{vice_poem}.

While earlier studies ~\citep{jiao2023chatgpt,sawicki2023bits, peng2023spoken}, have  evaluated LLM performance on poetry generation, machine translation and language learning. These studies were designed to analyse LLM performance on the downstream task rather than probe the model's inherent understanding of phonology. To this end, we devise three diagnostic tasks that benchmark various LLMs on fine-grained tasks that explicitly apply phonological concepts in English. Our benchmark, \textbf{\name{}}, comprises of three tasks - grapheme-to-phoneme (G2P) conversion, syllable counting and rhyme word generation, is shown in ~\autoref{fig:main}. G2P is a widely adopted task for speech recognition applications that leverages pronunciation information to produce phonemes. Additionally, syllable counting and rhyme word generation are educational exercises used to enhance early language skills in children \citep{bruck1995phonological}. Thus, \name{} provides a systematic framework to analyse the phonological skills of LLMs that inadvertently affects their performance in real-world applications (writing assistants, translators) ~\autoref{fig:main}.

We conduct a comprehensive study across 6 widely used LLMs, including three closed-source models - GPT-4 ~\citep{openai2023gpt4}, Claude-3-Sonnet ~\citep{claude}, GPT-3.5-Turbo ~\citep{chatgpt2022}, and three open-source models - \llama, Mistral-7B ~\citep{jiang2023mistral} and Mixtral-8X7B ~\citep{jiang2024mixtral}. Furthermore, we study the impact of word frequency, tokenization and sentence complexity across all three tasks. We find that overall LLMs exhibit surprisingly good performance on the three evaluated tasks while being trained solely on textual data. However, we observe a gap of 45\% between human performance and LLM performance on syllable counting whereas the gap is nearly 17\% for rhyme word generation. We also note that, based on our evaluation, no single model consistently outperforms the others across all evaluated tasks. For example, GPT-4 is the best model in Rhyme Generation while Claude-3-Sonnet outperforms GPT-4 by a large margin in Syllable Counting. This highlights that models are good at different phonological tasks and researchers should choose LLMs based on their performance on the core task related to the downstream applications.

Our main contributions are: 

\begin{enumerate}
    \item We introduce \name{}, a benchmark to evaluate the phonological awareness of LLMs on three diagnostic tasks in English - grapheme-to-phoneme conversion, syllable counting and rhyme word generation. \name{} offers 4k datapoints along various axes to holistically evaluate LLMs. 
    \item We benchmark six LLMs and study the impact of word frequency, tokenization and sentence complexity on LLM performance.  
    \item Our findings suggest that LLMs perform suboptimally in comparison to human. In addtion, we show that no single model is consistently good at all the tasks and researchers should choose LLMs that perform well on the tasks that affect the related downstream application.
\end{enumerate}

%% file: sections/related_work.tex
\section{Related Work}

\paragraph{Evaluation of LLM Linguistics.} \citet{hu2023promptbased} assesses the efficacy of prompting as a way of probing the model's metalinguistic ability, i.e., the ability to perform linguistic analyses given a natural language input. Contrary to our work, \citep{beguš2023large} presents qualitative case studies on phonology, syntax and semantics of GPT-4 by generating theoretical analyses of linguistic phenomenon on toy languages. Additionally,  \citep{basmov2023chatgpt} evaluates the performance of LLMs on linguistic inferences such as grammatically-specified entailments and monotonicity entailments using natural language inputs. More recently, ~\cite{peng2023spoken} proposes a multi-choice question answer dataset to evaluate the spoken language knowledge of LLMs. The proposed dataset comrpises of spoken language questions from an academic source (linguistics 101) whereas \name{} comprises of tasks that closely align with real-world applications.
 
 \paragraph{Grapheme-to-Phoneme.}
Many studies have been conducted on G2P conversion. Early works have proposed joint n-gram models~\citep{galescu2002pronunciation} and joint-sequence models ~\citep{bisani2008joint} for G2P conversion. Recent developments in G2P studies have shown
LSTM~\citep{toshniwal2016jointly} and Transformer~\citep{Yolchuyeva_2019} to be powerful G2P models. These models have access to various pronunciation dictionaries and are trained to explicitly accomplish the grapheme-phoneme conversion for speech applications like Automatic Speech Recognition(ASR) and Text-To-Speech (TTS) ~\citep{masumura2020phoneme}. Finally, ~\citep{g2pE2019} combines the CMUdict \footnote{http://www.speech.cs.cmu.edu/cgi-bin/cmudict} corpus with a neural-network model to convert graphemes to phonemes. However, previous works have not analysed the ability of LLMs as G2P models.

 \paragraph{Syllable Counting.}
Syllable counting is used for testing the phonological awareness of children at an early age ~\citep{bruck1995phonological, ukrainetz2011effects}. It is also useful in second language acquisition and commonly used as an educational tool. Additionally, counting of syllables is crucial for composing songs, poems and haikus ~\citep{tian2022zeroshot, henderson1965haiku, suzuki1985role}. Several works have studied the syllable structure of haikus and poetry in English, haikus are particularly popular for they 5-7-5 syllable structure ~\citep{henderson1965haiku, suzuki1985role}. Recently, ~\citep{tian2022zeroshot}, ~\citep{tian2023unsupervised} and ~\citep{ormazabal2022poelm} have utilized syllable counts as decoding contraints or metre descriptors to generate formatted sonnets, lyrics and poems. ~\cite{sun2023evaluating} evaluates the ability of LLMs to generate syllable-controlled texts (e.g. - Complete a sentence in 5 syllables). In this work,  we focus on evaluating the ability of LLMs to recognise and count the number of syllables in a sentence. 

 \paragraph{Rhyming and Rhymes.}
Rhyming words are pivotal in early age evaluation of language development in children and popularly used as an education tool to teach languages ~\citep{bruck1995phonological}. Additionally, rhyming words are critical components of creative writing tasks such as poetry and song writing ~\cite{caplan2014rhyme}. Prior works in automatic poetry and sonnet generation rely on external rhyming dictionaries sch as CMUdict to induce rhyme during generation ~\citep{tian2022zeroshot, tian2023unsupervised}. Recently, several works have studied LLM-generated poetry and creative artifacts with a focus on rhyme schemes, style and creativity ~\citep{sawicki2023bits}. However, previous works have not evaluated the ability of LLMs to generate corresponding rhyming words as a phonological skill assessment.  

%% file: sections/benchmark.tex
\section{\name{}}
Here, we present our benchmark, \name{}, that evaluates the English phonological skills of LLMs across three tasks: Grapheme-to-Phoneme Conversion (\S \ref{sec:g2p}), Syllable Counting (\S \ref{sec:syllable}), and Rhyme Word Generation (\S \ref{sec:rhyme}). We provide an overview of these tasks and their applications in various downstream tasks in \autoref{fig:main}. An overview of these tasks, along with their significance in a variety of downstream applications, is provided in \autoref{fig:main}. Starting with Grapheme-to-Phoneme Conversion, this task is pivotal for enhancing speech-related applications, including Automatic Speech Recognition and Text-to-Speech systems, by ensuring accurate phonetic interpretations of text. Furthermore, the incorporation of phonemic and phonetic transcriptions into machine translation models introduces additional layers of understanding through phonetic embeddings, thereby improving translations ~\citep{liu-etal-2019-robust, chakravarthi2019multilingual}. In addition, Syllable Counting and Rhyme Word Generation are traditionally popular as educational tools. Beyond facilitating language learning, they are essential in the creative processes involved in poetry and song composition, illustrating their versatility. Overall, we find that the tasks curated in \name{} shed light on a wide range of real-world applications for LLMs, and benchmarking LLM performance on these tasks can advace research in LLM development for improved linguistic and creative generations. We provide the dataset statistics in ~\autoref{tab:dataset_statistics} and outline the experimental methodology in \S ~\ref{sec:g2p}-~\ref{sec:rhyme}. 

\input{sections/stats_table}

\subsection{Grapheme-to-Phoneme Generation}
\label{sec:g2p}
\paragraph{Task Description.} The grapheme-to-phoneme (G2P) task involves converting the orthographic representation of a word into its phonemic representation.  Writing systems often do not have a one-to-one mapping with spoken forms in English. For example, the element \textit{o} in `olive(/\textscripta/)' differs phonetically from the \textit{o} in `rose(/o\textupsilon/)'. Therefore, inferring phonetic transcriptions solely from orthography is challenging. \textbf{In this task, we prompt LLMs with a word to predict the correct phonemic form in International Phonetic Alphabet (IPA).} 

\paragraph{Dataset and Evaluation Metric.} We sample 3,000 words and their corresponding pronunciations from the SIGMORPHON 2021 G2P task ~\citep{ashby2021results} for American English to curate our dataset. Then, we probe the LLMs with the prompt (zero-shot) shown in ~\autoref{table:prompts} and report the accuracy for each model. Accuracy is the percentage of words whose predicted phoneme sequences were identical to the gold references. 

\paragraph{Baseline.} We report the performance of the G2P ~\citep{g2pE2019} library on our dataset as the baseline. This library is a combination of a dictionary look-up and a neural network and serves as a reasonable baseline with access to pronunciation information. 

\subsection{Syllable Counting}
\label{sec:syllable}
\paragraph{Task Description.} In syllable counting, individuals must identify the vowel peaks and the consonants that may precede or follow these peaks to determine the number of syllables in a word ~\citep{bruck1995phonological}. Syllable counting is an educational tool used to improve a child's phonemic awareness. Therefore, assessing the performance of LLMs can provide insights into the model's phonemic awareness. Here, \textbf{we prompt LLMs with a sentence to count the number of syllables} in the sentence. 

\paragraph{Dataset and Evaluation Metric.}
Following ~\citep{sun2023evaluating}, we curate 1,000 sentences from the datasets of Romance Books and Reddit Short Stories \footnote{https://www.kaggle.com/datasets/trevordu/reddit-short-stories}. After that, we employ a grapheme-to-phoneme library to count the syllables in the sentences ~\citep{g2pE2019}. Finally, we probe LLMs with the prompt shown in Table~\ref{table:prompts} and report the accuracy for each model. Accuracy is the percentage of sentences whose predicted syllable counts are correct. 

\paragraph{Baseline.} 
We implement a naive approach by counting the number of vowels (a,e,i,o,u) in the sentence (text) as a baseline. Additionally, we ask an annotator with graduate-level engineering education and working proficiency in English (US) to perform the task and treat it as the human performance baseline.

\subsection{Rhyme Generation}
\label{sec:rhyme}
\paragraph{Task Description.} In this task, we aim to analyse the LLMs' capability to generate correct rhyming words for a given word. Rhyme words are a crucial aspect of creative writing and rely on pronunciation knowledge. Since, LLMs are used popularly as creative writing assitants, LLM performance on this task provides insights about creative generation.  

\paragraph{Dataset and Evaluation Metric.} We collect 300 words from the Spelling Bee Study List ~\citep{maguire2006american} and the Google One Trillion corpus \footnote{\url{https://github.com/first20hours/google-10000-english}} to curate our dataset. We retrieve all the rhyming words (slant and strict rhymes) for a given word from an online rhyming dictionary, WordHippo\footnote{\url{https://www.wordhippo.com/}} and treat these as the gold reference. We obtain an average of 1200 rhyming words for a word to ensure a good coverage in our gold references.  Similar to ~\autoref{sec:g2p}, we categorized the dataset into two segments - high-frequency words and low-frequency words by using WIMBD ~\citep{elazar2024whats} index. Finally, we probe LLMs with the prompt shown in~\autoref{table:prompts} and report the Success Rate (SR). We compute the word-specific success rate as the number of generated rhyming candidates that belong to the ground-truth set of rhyming words. SR is the average success rate for all the words.

\paragraph{Baseline.} We present human performance as a baseline. Two human annotators with college-level education and native proficiency in English (US) performed the task. All human annotators were paid an \$18 per hour and we spent \$100 to acquire all annotations. We finetune a LLaMa-2-13B-Chat model (2 epochs, 2e-6 learning rate) on the common words and evaluate its performance on the rare words as a task-specific baseline.

%% file: sections/stats_table.tex
\begin{table}[h]
\centering
\begin{tabular}{cc}
\hline
\textbf{Grapheme-to-Phoneme}&  \textbf{Number}    \\\hline
Number of Words & 3K \\
High Frequency/Low Frequency & 1K/2k  \\
Whole-Word/Split-Word (High Frequency) & 700/200  \\\hline    
\textbf{Syllable Counting} &  \\\hline
Number of Sentences & 1K \\
Simple/Complex & 740/260 \\\hline
\textbf{Rhyme Word Generation}                                                   &       \\\hline
Number of Words & 300 \\
Common/Rare & 200/100  \\\hline
\end{tabular}
\caption{Dataset Statistics of the proposed \name{}.}
\label{tab:dataset_statistics}
\end{table}

%% file: sections/experimental_results.tex
\section{Experimental Results}
\label{sec:results}
\begin{figure}
  \begin{center}
    \includegraphics[width=0.8\textwidth]{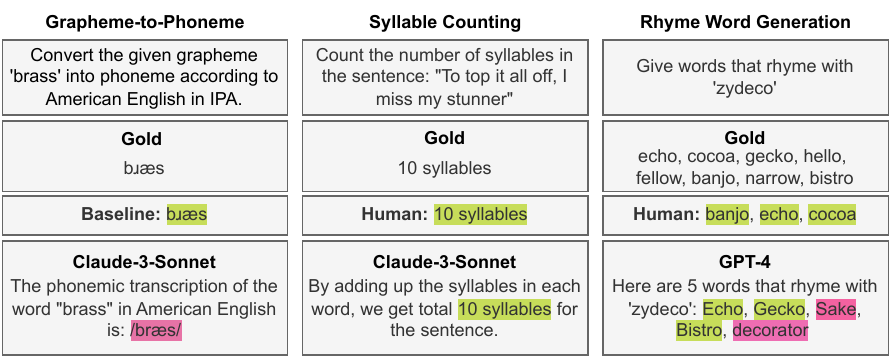}
  \end{center}
  \caption{\small{Samples from all three tasks in \name{}}}
  \label{fig:data_samples}
\end{figure}

Here, we aim to benchmark the phonological skills of LLMs across the three tasks - G2P (\S ~\ref{sec:g2p_results}), Syllable Counting (\S ~\ref{sec:syllable_results}) and Rhyme Word Generation (\S ~\ref{sec:rhyme_results}). We further study the impact of word frequency in the pretraining dataset, tokenization strategy, and sentence complexity on LLM performance across all the tasks.~\autoref{fig:data_samples} shows human as well as model responses on the three tasks along with the gold references. We prompt all models in a zero-shot setting.

\subsection{Grapheme-to-Phoneme}
\label{sec:g2p_results}
\paragraph{Overall Model Performance.}
Our results in~\autoref{tab:g2p_freq_results} show that LLMs are worse than the phonologically trained baseline, G2P by $\sim$10\% with Claude-3-Sonnet and GPT-4 achieving the highest performance. We observe that the best performing open-source model (with <15B params), \llama is $\sim$34\% behind GPT-4. We provide qualitative examples in~\autoref{fig:phoneme_egs}.

\paragraph{Frequently Used Words.} 
Usually, words that are more commonly used in real-world scenarios are more prevalent in the training dataset. Highly frequent words may also afford LLMs greater opportunities to learn their correct pronunciations during training as shown in numerical reasoning tasks ~\citep{razeghi2022impact}. To explore LLM performance on different frequency words, we categorized the dataset into two segments - high-frequency words and low-frequency words. We count the frequency of all the words in our dataset using the index in WIMBD ~\citep{elazar2024whats} \footnote{WIMBD returns the number of documents where a word occurs in the C4 ~\citep{raffel2023exploring} dataset. We use this as a proxy for occurrence frequency.}. Our dataset comprises 1,000 high-frequency words with more than 10M occurrences and 2,000 low-frequency words that occur less than 1M in pretraining corpora. Results in ~\autoref{tab:g2p_freq_results} indicate that models (including the baseline) generally have higher performance on high-frequency words than on low-frequency words. This could be attributed to wider availability of pronunciation information for high-frequency words in the training dataset as well as online dictionaries. However, our baseline model still outperforms the best LLM (Claude-3-Sonnet) on both high-frequency and low-frequency words highlighting the need for curating better phonology-rich datasets.

\begin{table}[h]
    \centering
    \begin{tabular}{lccc}
    \hline
\textbf{Model} & \textbf{High Frequency} & \textbf{Low Frequency} \\
\hline
&Open-Sourced Models&\\
\hline
\llama & 18.0 & 12.6 & \\
Mistral-7B-Instruct & 5.3 & 2.4\\
Mixtral-8X7B-Instruct & 22.0 & 18.1\\
\hline
&Closed-Sourced Models&\\
\hline
GPT-3.5-Turbo & 47.6 & 34.4 \\
GPT-4 & 51.9 & 38.1 \\
Claude-3-Sonnet & \colorbox{junebud}{52.7} & \colorbox{junebud}{40.2} \\
\hline
&Baseline&\\
\hline
G2P~\citep{g2pE2019} & 62.4 & 52.8 \\
\hline
\end{tabular}
\caption{Results for grapheme-to-phoneme conversion by LLMs. We report the \textit{Accuracy} as the percentage of correct phonemes generated by each LLM. \textit{High Frequency} words are words that occur 100 times more than \textit{Low Frequency} words in pretraining corpora.}
\label{tab:g2p_freq_results}
\end{table}

\begin{table}[h]
    \centering
    \begin{tabular}{lccc}
    \hline
\textbf{Model} & \textbf{Whole-Word} & \textbf{Split-Word} \\
\hline
&Open-Sourced Models&\\
\hline
\llama & 22.7 & 15.9 \\
Mistral-7B-Instruct & 8.0 & 4.0 \\
Mixtral-8X7B-Instruct & 27.9 & 19.5\\
\hline
&Closed-Sourced Models&\\
\hline
GPT-3.5-Turbo & 53.1 & 45.1 \\
GPT-4 & 58.0 & \colorbox{junebud}{49.2} \\
Claude-3-Sonnet & \colorbox{junebud}{64.8} & 47.3 \\
\hline
\end{tabular}
\caption{Results for grapheme-to-phoneme conversion by LLMs. We report the \textit{accuracy} as the percentage of correct phonemes generated by each LLM. \textit{Whole-Words} are preserved during tokenization by OpenAI tokenizer while {Split-Words} are split into sub-word tokens. We report LLM performance on the high-frequency words.}
\label{tab:g2p_token_results}
\end{table}

\paragraph{Tokenization in LLMs.}
Tokenization, refers to the division of input sequence of bytes into discrete tokens. LLMs primarily use Byte Pair Encoding (BPE)~\cite{shibata1999byte} or its variants ~\citep{kudo2018sentencepiece} for tokenization, leading to a generative process predominantly based on subword generation. We argue that this tokenization process can result in the loss of a word's phonological structure, which in turn, may impede the phoneme generation process. To investigate this further, we divided high-frequency words into two categories: whole-word tokens and split-word tokens. We utilize the tiktoken library, which provides open access to the tokenizer used in OpenAI's models, for word tokenization and subsequent data segmentation. Our analysis revealed that 30\% of the high frequency words were tokenized as whole words by the OpenAI tokenizer. Results in~\autoref{tab:g2p_token_results} show that LLMs achieve higher accuracy with whole-word tokens compared to split-word tokens. We also observe that GPT-4 outperforms Claude-3-Sonnet on split-words. This highlights that subword tokenization may lead to loss of phonological information that eventually affects the phonological skills of LLMs.

\begin{figure}
  \begin{center}
    \includegraphics[width=0.7\textwidth]{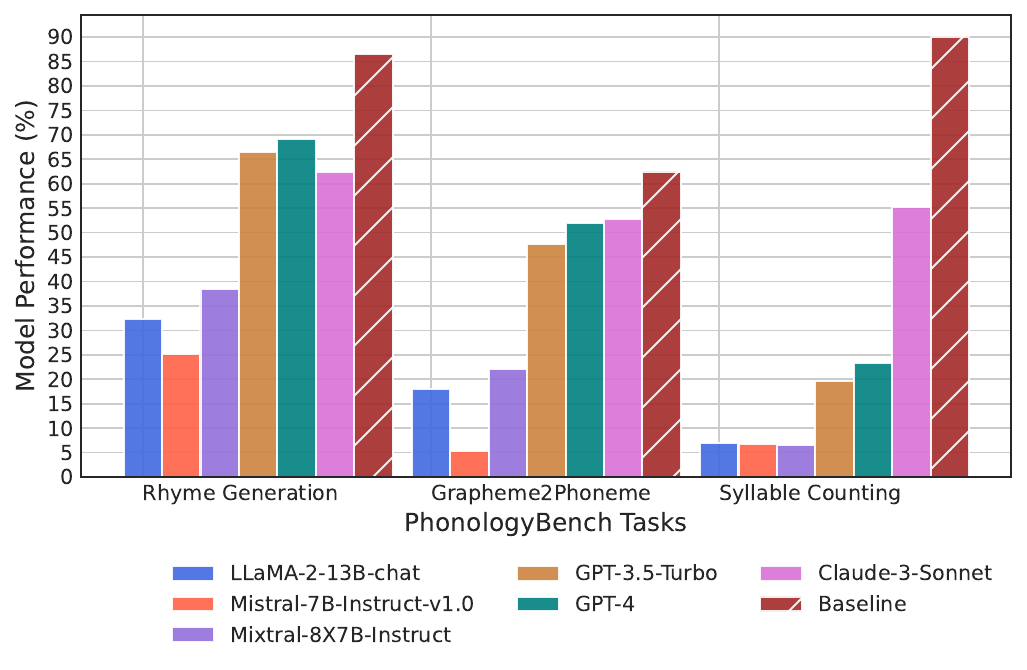}
  \end{center}
  \caption{\small{Performance of the 6 evaluated models on \name{}. We consider human performance on Rhyme Word Generation and Syllable Counting as the baseline. For G2P, we consider a dictionary look-up based approach as the baseline. }}
  \label{fig:main_results}
\end{figure}

\subsection{Syllable Counting}
\label{sec:syllable_results}

\paragraph{Overall Model Performance.} We find that Claude-3-Sonnet achieves the best performance of 55\% which is far behind the human baseline at 90\%. All the closed-source models beat the vowel baseline. Open-source models like \llama and Mistral-7B perform worse than the vowel baseline indicating that these models do not have an innate understanding of syllable structure and their relation with vowels. Surprisingly, GPT-4 falls behind Claude-3-Sonnet by $\sim$22\% while being at-par in G2P task. We provide qualitative examples in~\autoref{fig:syllable_egs}.

\paragraph{Sentence Complexity and Sentence Length.}
 A syntactically complex sentence is hard to comprehend due to the increased cognitive load required to grasp the syntax and semantics of the sentence ~\citep{mikk2008sentence}. We posit that LLMs face similar difficulties in performing reasoning tasks like syllable counting over syntactically complex sentences. Therefore, we group our dataset into two categories : simple and complex sentences \footnote{We overload the term 'complex' sentence to refer to sentences with more than one clause to simplify our analysis}. We utilize the spacy library \footnote{https://spacy.io/} to identify the number of clauses in a sentence. Our dataset comprises 74\% simple sentences and 26\% complex sentences. Our results are shown in~\autoref{tab:syllable_results}. We observe that with increasing sentence complexity model performance drops significantly. Usually, syntactically complex sentences are positively correlated with sentence length, hence, we show the performance of LLMs on differing sentence lengths in Appendix ~\ref{sec:sentence_length}. This highlights that LLMs can leverage the statistical correlations they learned from orthography to accomplish simple tasks. However, they tend to degrade in performance when faced with complex task and increased ambiguity. This shows that LLMs are accomplishing phonology-rich tasks by utilizing side evidence from the training data instead of phonological concepts and reasoning as humans do. To our surprise, for Claude-3-Sonnet the performance is higher for complex sentences than simple sentences. This could be due to training data bias where models have seen many complex and longer sentences found in literature, academic papers, and professional communications ~\citep{elazar2024whats}. Moreover, Claude-3 models have shown near perfect performance in long context evaluations indicating that they are good at processing longer inputs ~\citep{claude}. 

\begin{table}[t]
    \centering
    \begin{tabular}{lccc}
    \hline
\textbf{Model } & \textbf{Overall} & \textbf{Simple} & \textbf{Complex}   \\\hline
&Open-Sourced Models&\\
\hline
\llama & 6.9 & 7.5 & 5.1\\
Mistral-7B-Instruct & 6.8 & 7.0 & 6.2\\
Mixtral-8X7B-Instruct & 6.6 & 7.0 & 5.5\\
\hline
&Closed-Sourced Models&\\
\hline
GPT-3.5-Turbo & 19.6 & 20.2 & 14.0\\
GPT-4 & 23.3 & 24.2 & 15.3\\
Claude-3-Sonnet & \colorbox{junebud}{55.3} & \colorbox{junebud}{53.4} & \colorbox{junebud}{65.0}\\
\hline
&Baseline&\\
\hline
Vowel Baseline & 12.1 & 12.4 & 12.1 \\
Human & 90.0 & 93.3 & 86.7 \\
\hline
\end{tabular}
\caption{Results for syllable counting by LLMs. Here, we report the \textit{Accuracy} as the percentage of correct syllable counts predicted by LLMs. \textit{Simple} sentences contain only one clause while \textit{Complex} sentences contain more than one clause. }
\label{tab:syllable_results}
\end{table}
\subsection{Rhyme Word Generation}
\paragraph{Overall Model Performance.} We find that GPT-4 achieves the best overall performance of 57.6\%. In comparison, open-source models such as LLaMA and Mixtral-8X7B lag behind their closed-source counterparts, with Mixtral-8X7B securing the highest success rate among them at 27.9\%. Overall, humans beat all the evaluated models by 17\%. Intriguingly, supervised fine-tuning of a \llama-Chat model does not yield any performance enhancements. This suggests that additional training focused on orthography does not contribute to improving the model's capabilities in generating rhyme. We provide qualitative examples in~\autoref{fig:rhyme_egs}.

\paragraph{Impact of Word Frequency} 
Similar to \S~\ref{sec:g2p}, we study LLM performance on words with varying frequency of usage. Our dataset comprises 200 common words with more than 10M appearances and 100 rare words that occur less than 10K in in the proxy pretraining dataset (C4). Rare words are less commonly used by English speakers and their pronunciations may not be widely known though available in large datasets. Results in~\autoref{tab:rhyme_results} indicate that LLMs exhibit higher accuracy in rhyme generation for common words compared to their rare counterparts. This could be due to less exposure to rare words and their pronunciations during training and highlights the importance of phonological information for improved LLM performance.

\label{sec:rhyme_results}
\begin{table}[t]
    \centering
    \begin{tabular}{lcc}
    \hline
\textbf{Model}  & \textbf{Common Words} & \textbf{Rare Words}  \\\hline
&Open-Sourced Models&\\
\hline
\llama &  32.4 & 15.6\\
Mistral-7B-Instruct-v0.1 &  25.1 &8.3  \\
Mixtral-8X7B-Instruct &  38.4  & 17.5\\
\hline 
&Closed-Sourced Models&\\
\hline
GPT-3.5-Turbo & 66.5 & 42.7  \\
GPT-4 & \colorbox{junebud}{69.1} & \colorbox{junebud}{46.1}\\
Claude-3-Sonnet & 62.4 &  39.6 \\
\hline
LLaMA-2-13B (SFT) & 15.8 & 15.8 \\
\hline
&Baseline&\\
\hline
Human & 86.4 & 60.4 \\
\hline
\end{tabular}
\caption{Results for rhyme word generation by LLMs. Here, we report the \textit{Success Rate} as the correct rhyming words out the five generated per word by LLM. \textit{Common Words} are words that occur 1000 times more than \textit{Rare Words} in pretraining corpora.}
\label{tab:rhyme_results}
\end{table}

\subsection{Discussion}
In summary, our findings suggest that despite being trained on orthographic form \textbf{LLMs perform surprisingly well on English phonological tasks}. This could be attributed to the imperfect associations between text and speech learned by LLMs during to their training. Additionally, 
in~\autoref{fig:main_results} we show that despite notable performance on different tasks, \textbf{LLMs fall behind relevant baselines across all tasks.} We find that the evaluated open-source models are consistently worse than the closed-source models. However, we also observe that \textbf{no single model consistently} outperforms the others in these tasks. For example, while \llama outperforms Mistral-Instruct-7B on rhyme generation and G2P, it has the same performance on Syllable Counting. This indicates that higher performance in one task does not signify higher performance across all tasks, thus researchers should choose LLMs based on their performance on the core task for related downstream applications.

%% file: sections/conclusion.tex
\section{Conclusion}
In this work, we present, \name{} that consists of 4k data samples to evaluate the phonological skills of LLMs across G2P conversion, syllable counting and rhyme word generation in English. We show that LLMs fail to outperform human performance in syllable counting and rhyme word generation. Our findings thus highlight that there is scope of improving LLM performance on these core phonological task and encourages future research in training phonologically aware models. A straightforward approach to improving model performance across these phonological tasks is to add more phonological data during pre-training. Prior work by \citet{liu-etal-2019-robust} have shown the efficacy of joint textual and phonetic embedding in neural machine translation, thus, future work can focus on augmenting LLMs with phonetic representations. Overall, we show that no single model consistently performs well on all the tasks. Thus, we encourage researchers to develop downstream applications that rely on these phonological tasks by carefully selecting LLMs that perform well on the particular task. 

%% file: sections/appendix.tex
\section{Limitations}
Our work analyses LLM performance on various phonological tasks. The limitations of our work are two-fold, including concept coverage and data quality.
While we present a comprehensive study on the various facets of LLM performance on G2P, syllable counting and rhyme word generation, our work only covers a limited portion of the variety of phonological phenomenon such as blending, homonyms and homographs. Additionally, our work focuses on American English and does not extend to various dialects. Future studies can focus on multilingual phonological phenomenon as well as LLM performance on different dialects. We encourage research that addresses these various languages, dialects and phenomenon to further the understanding of LLM language skills. Finally, our benchmark though extensive is limited by the quality of the curated gold references. For example, despite a large coverage of rhyming words we cannot be certain that our gold references are exhaustive. Similarly, LLMs can process syllable counts differently for clinical data or scientific communications that is not covered in \name{}. Therefore, we encourage further contributions from the research community to help develop high-quality evaluation sets.

\section{Syllable Counting at Varying Sentence Lengths}
\label{sec:sentence_length}
\begin{figure}
  \begin{center}
    \includegraphics[width=0.7\textwidth]{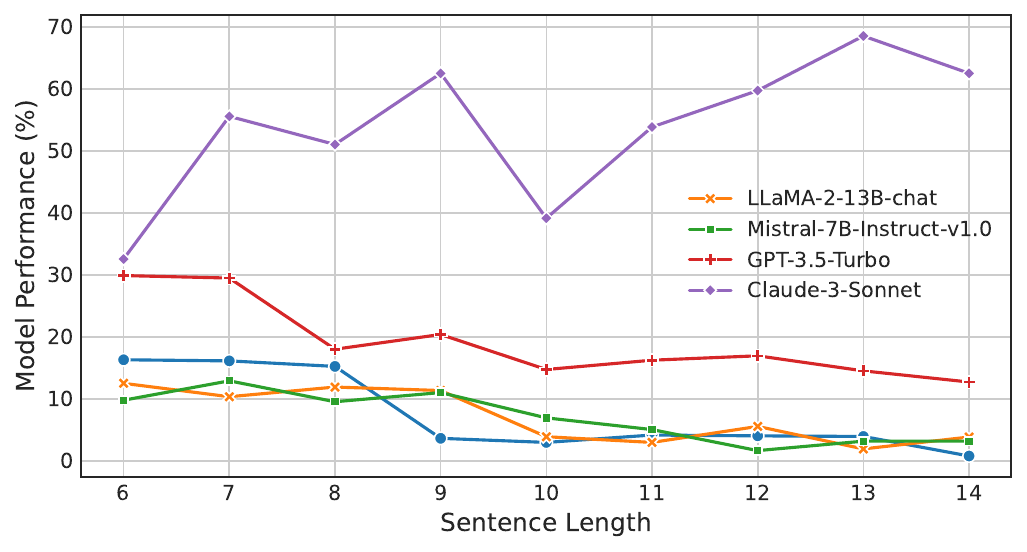}
  \end{center}
  \caption{\small{Performance of the 4 models on syllable counting at varying sentence lengths. }}
  \label{fig:syllable_counting}
\end{figure}

In this section, we study the impact of sentence length on LLMs on the syllable counting task. ~\autoref{fig:syllable_counting} shows that overall models have higher accuracy for shorter sentence (length=6) than longer sentences(length=13). This highlights the relationship between sentence complexity and sentence length with similar trends as \S~\ref{sec:syllable_results}.
\begin{table*}[h]
\centering
\footnotesize
\begin{tabular}{ll}
\toprule
\textbf{Task} &\textbf{Prompt}  \\
\toprule
Grapheme-to-Phoneme & \begin{tabular}[l]{@{}l@{}}
You are an expert in American English phonology, phonetics and morphology. \\ In this task, you are required to map a sequence of graphemes -characters \\representing a word to a transcription of that word's pronunciation (phonemes).\\ If you cannot complete the task, respond with - NONE.  \\
\\
Generate the phoneme for the grapheme :'<input>'
\end{tabular}\\
\midrule
 Rhyme Word Generation & \begin{tabular}[l]{@{}l@{}}Rhyming words are words that have the same ending sound.\\ In simpler terms, it can be defined as the repetition \\ of similar ending sounds.\\ \\ Give 5 words that rhyme with '<input>'. \end{tabular}\\  
 \midrule
 Syllable Counting & \begin{tabular}[l]{@{}l@{}}
 Syllable is a unit of pronunciation having one vowel sound, with or without \\ surrounding consonants, forming the whole or a part of a word.\\
 
 \\Count the number of syllables in the given sentence :'<input>' \end{tabular}\\

\bottomrule
\end{tabular}
\caption{Example Prompts for Tasks. Each prompt is prefixed with general information about the task and concepts.}
\label{table:prompts}
\end{table*}
\section{Qualitative Examples}
In this section, we present the qualitative examples to show the varied model responses on different tasks of our dataset ~\autoref{fig:phoneme_egs} - ~\ref{fig:rhyme_egs}. 

\input{sections/qualitative}
\input{sections/qualitative_transcription}
\input{sections/qualitative_rhyme}

%% file: sections/qualitative.tex
\begin{table*}[h]
\begin{tabular}{ccc}

\begin{minipage}[h]{0.5\textwidth}
\vspace*{0pt}
\fontsize{9.0pt}{\baselineskip}\selectfont
\linespread{0.9}\selectfont
\begin{mybody}
\textbf{Instruction:} Count syllables in this sentence: "She meets him randomly in the woods at his family's cabin."\\
\textbf{Ground Truth Count:} 
16
\end{mybody}
\begin{mybody}
\textbf{GPT-4:} Counting the syllables in the phrase "She meets him randomly in the woods at his family's cabin," we have:\\

She: 1 syllable \\
meets: 1 syllable \\
him: 1 syllable \\
randomly: 3 syllables \\
in: 1 syllable \\
the: 1 syllable \\
woods: 1 syllable \\
at: 1 syllable \\
his: 1 syllable \\
family's: 3 syllables \\
cabin: 2 syllables \\
So, the total is 16 syllables in this phrase. \\

\end{mybody}
\begin{mybody}
\textbf{Claude-3-Sonnet:} To count the number of syllables in the given sentence, we need to break it down into individual words and count the syllables in each word. Here's the breakdown: \\

"She" - 1 syllable \\
"meets" - 1 syllable \\
"him" - 1 syllable \\
"ran-dom-ly" - 3 syllables \\
"in" - 1 syllable \\
"the" - 1 syllable \\
"woods" - 1 syllable \\
"at" - 1 syllable \\
"his" - 1 syllable \\
"fam-i-ly's" - 3 syllables \\
"cab-in" - 2 syllables \\
 
Adding up the syllables for each word, we get: 16 syllables \\
\end{mybody}
\begin{mybody}
\textbf{Mistral-7b-instruct:} 
The number of syllables in the given sentence is 12.\\
\end{mybody}

\begin{mybody}
\textbf{Llama-2-Chat:} Sure! Here's the number of syllables in each word of the sentence: \\

1. She - 1 syllable \\
2. meets - 2 syllables \\
3. him - 1 syllable \\
4. randomly - 3 syllables \\
5. in - 1 syllable \\
6. the - 1 syllable \\
7. woods - 2 syllables \\
8. at - 1 syllable \\
9. his - 1 syllable \\
10. family's - 2 syllables \\
11. cabin - 1 syllable \\

Total number of syllables: 11 \\
\end{mybody}
\end{minipage} &

\begin{minipage}[h]{0.5\textwidth}
\vspace*{0pt}
\fontsize{9.0pt}{\baselineskip}\selectfont
\linespread{0.9}\selectfont
\begin{mybody}
\textbf{Instruction:} Count syllables in this sentence: "To top it all off, I miss my Stunner"\\
\hfill\break
\textbf{Ground Truth Count:} 
10
\end{mybody}
\begin{mybody}
\textbf{GPT-4:} Counting the syllables in the phrase "To top it all off, I miss my Stunner," we have: \\
To: 1 syllable \\
top: 1 syllable \\
it: 1 syllable\\
all: 1 syllable\\
off: 1 syllable\\
I: 1 syllable\\
miss: 1 syllable\\
my: 1 syllable\\
Stun-ner: 2 syllables\\
The total is 10 syllables in this phrase. \\

\end{mybody}
\begin{mybody}
\textbf{Claude-3-Sonnet:} To count the number of syllables in the given sentence, we need to break it down into individual words and then count the syllables in each word.\\

The sentence is: "To top it all off, I miss my Stunner."\\

Here's the breakdown:\\

1. "To" - 1 syllable\\
2. "top" - 1 syllable\\
3. "it" - 1 syllable\\
4. "all" - 1 syllable\\
5. "off" - 1 syllable\\
6. "I" - 1 syllable\\
7. "miss" - 1 syllable\\
8. "my" - 1 syllable\\
9. "Stun-ner" - 2 syllables\\

By adding up the syllables in each word, we get:\\

1 + 1 + 1 + 1 + 1 + 1 + 1 + 1 + 2 = 10 syllables\\
Therefore, the given sentence "To top it all off, I miss my Stunner." has 10 syllables.\\

\end{mybody}
\begin{mybody}
\textbf{Mistral-7b-instruct:} 
The number of syllables in the given sentence is 10.\\

\end{mybody}

\begin{mybody}
\textbf{Llama-2-Chat:} Sure! Here's the syllable count for the sentence "To top it all off, I miss my Stunner": \\
1. To\\
2. top\\
3. it\\
4. all\\
5. off\\
6. I\\
7. miss\\
8. my\\
9. Stunner\\
There are 9 syllables in the sentence.\\

\end{mybody}

\end{minipage} &

\end{tabular}
\captionof{figure}{We provide responses of GPT-4, Claude-3-Sonnet, \llama and Mistral-Instruct-7B on two samples from the Syllable Counting task.}
\label{fig:syllable_egs}
\end{table*}

%% file: sections/qualitative_transcription.tex
\begin{table*}[h]
\begin{tabular}{ccc}

\begin{minipage}[h]{0.5\textwidth}
\vspace*{0pt}
\fontsize{9.0pt}{\baselineskip}\selectfont
\linespread{0.9}\selectfont
\begin{mybody}
\textbf{Instruction:} Give the phonetic transcription of `Brass’ in IPA"\\
\textbf{Ground Truth:} /b r æ s/ 

\end{mybody}
\begin{mybody}
\textbf{GPT-4:} Word :  The phonetic transcription for "brass" in IPA in American English is /bræs/.\\

\end{mybody}
\begin{mybody}
\textbf{Claude-3-Sonnet:} The phonemic transcription of the word "brass" in American English is: /bræs/\\

\end{mybody}
\begin{mybody}
\textbf{Mistral-7b-instruct:} The phoneme transcription for the grapheme sequence "brass" is /bræs/ \\

\end{mybody}

\begin{mybody}
\textbf{Llama-2-Chat:} 
Sure, I can help you with that! The word "brass" has the following graphemes: b-r-a-s-s And its transcription in phonemes would be: /bræs/ \\
\end{mybody}

\end{minipage} &

\begin{minipage}[h]{0.5\textwidth}
\vspace*{0pt}
\fontsize{9.0pt}{\baselineskip}\selectfont
\linespread{0.9}\selectfont
\begin{mybody}
\textbf{Instruction:} Give the phonetic transcription of `Oath’ in IPA"\\
\textbf{Ground Truth: \textipa{/o\textupsilon \texttheta/}} 

\end{mybody}
\begin{mybody} 
\textbf{ GPT-4:} The phonetic transcription for "oath" in IPA in American English \textipa{/o\textupsilon \texttheta/}\\

\end{mybody}
\begin{mybody}
\textbf{Claude-3-Sonnet:} The phonemic transcription of the word "oath" in American English is:   \textipa{/o \textupsilon \texttheta/} \\

\end{mybody}
\begin{mybody}
\textbf{Mistral-7b-instruct:} The phoneme transcription for the word "oath" is \textipa{/o:\texttheta/} \\

\end{mybody}

\begin{mybody}
\textbf{Llama-2-Chat:} 
Sure, I can help you with that! The word "oath" has a pronunciation of \textipa{/O\texttheta/}. So, the phoneme for the grapheme "oath" is \textipa{/O/}.\\
\end{mybody}

\end{minipage}&

\end{tabular}
\captionof{figure}{We provide responses of GPT-4, Claude-3-Sonnet, \llama and Mistral-Instruct-7B on two samples from the G2P task.}
\label{fig:phoneme_egs}
\end{table*}

%% file: sections/qualitative_rhyme.tex
\begin{table*}[h]
\begin{tabular}{ccc}

\begin{minipage}[h]{0.5\textwidth}
\vspace*{0pt}
\fontsize{9.0pt}{\baselineskip}\selectfont
\linespread{0.9}\selectfont
\begin{mybody}
\textbf{Instruction:} Give 5 words that rhyme with submitted."\\
\textbf{Ground Truth:} Committed, Admitted, Acquitted, Outwitted,
Transmitted, Permitted, Uplifted,
Omitted.

\end{mybody}
\begin{mybody}
\textbf{GPT-4:} Word :  submitted \\
1. Admitted\\
2. Committed\\
3. Permitted\\
4. Omitted\\
5. Transmitted\\

\end{mybody}
\begin{mybody}
\textbf{Claude-3-Sonnet:} Here are 5 words that rhyme with "submitted":\\

1. Committed\\
2. Omitted\\
3. Transmitted\\
4. Outwitted\\
5. Permitted\\

\end{mybody}
\begin{mybody}
\textbf{Mistral-7b-instruct:} Some words that rhyme with "submitted" are:\\
1. Consumed\\
2. Exhumed\\
3. Jumped \\
4. Lumbered\\
5. Rumpus\\

\end{mybody}

\begin{mybody}
\textbf{Llama-2-Chat:} 
Sure! Here are five words that rhyme with "submitted":\\

1. submitted\\
2. committed\\
3. submitted\\
4. neglected\\
5. invested\\

\end{mybody}

\end{minipage} &

\begin{minipage}[h]{0.5\textwidth}
\vspace*{0pt}
\fontsize{9.0pt}{\baselineskip}\selectfont
\linespread{0.9}\selectfont
\begin{mybody}
\textbf{Instruction:} Give 5 words that rhyme with trademarks."\\
\textbf{Ground Truth:} Remarks, Landmarks, Watermarks, Marks, Barks, Sharks, Hallmarks, Parks, Sparks, Benchmarks, Larks, Monarchs.
\end{mybody}
\begin{mybody}
\textbf{ GPT-4:}Word :  trademarks\\
1. Parks\\
2. Sparks\\
3. Sharks\\
4. Larks\\
5. Marks\\

\end{mybody}
\begin{mybody}
\textbf{Claude-3-Sonnet:} Here are 5 words that rhyme with "trademarks":\\

1. Spacebarks\\
2. Masquerades\\
3. Charades\\
4. Crusades\\
5. Blockades\\

\end{mybody}
\begin{mybody}
\textbf{Mistral-7b-instruct:} Sure are five words that rhyme with "trademarks":\\
1. Sparklines\\
2. Landmarks\\
3. Darkening\\
4. Outsmarting\\
5. Repairs\\

\end{mybody}

\begin{mybody}
\textbf{Llama-2-Chat:} 
Sure, here are five words that rhyme with "trademarks":\\
1. brands\\
2. marks\\
3. sparks\\
4. sharks\\
5. arks\\

\end{mybody}

\end{minipage}\\

\end{tabular}
\captionof{figure}{We provide responses of GPT-4, Claude-3-Sonnet, \llama and Mistral-Instruct-7B on two samples from the Rhyme Word Generation task.}
\label{fig:rhyme_egs}
\end{table*}

%% file: main.bbl
\begin{thebibliography}{49}
\providecommand{\natexlab}[1]{#1}
\providecommand{\url}[1]{\texttt{#1}}
\expandafter\ifx\csname urlstyle\endcsname\relax
  \providecommand{\doi}[1]{doi: #1}\else
  \providecommand{\doi}{doi: \begingroup \urlstyle{rm}\Url}\fi

\bibitem[Anthropic(2024)]{claude}
Anthropic.
\newblock Introducing the next generation of claude.
\newblock \url{https://www.anthropic.com/news/claude-3-family}, 2024.

\bibitem[Ashby et~al.(2021)Ashby, Bartley, Clematide, Del~Signore, Gibson, Gorman, Lee-Sikka, Makarov, Malanoski, Miller, et~al.]{ashby2021results}
Lucas~FE Ashby, Travis~M Bartley, Simon Clematide, Luca Del~Signore, Cameron Gibson, Kyle Gorman, Yeonju Lee-Sikka, Peter Makarov, Aidan Malanoski, Sean Miller, et~al.
\newblock Results of the second sigmorphon shared task on multilingual grapheme-to-phoneme conversion.
\newblock In \emph{Proceedings of the 18th SIGMORPHON Workshop on Computational Research in Phonetics, Phonology, and Morphology}, pp.\  115--125, 2021.

\bibitem[Bang et~al.(2023)Bang, Cahyawijaya, Lee, Dai, Su, Wilie, Lovenia, Ji, Yu, Chung, Do, Xu, and Fung]{bang2023multitask}
Yejin Bang, Samuel Cahyawijaya, Nayeon Lee, Wenliang Dai, Dan Su, Bryan Wilie, Holy Lovenia, Ziwei Ji, Tiezheng Yu, Willy Chung, Quyet~V. Do, Yan Xu, and Pascale Fung.
\newblock A multitask, multilingual, multimodal evaluation of chatgpt on reasoning, hallucination, and interactivity, 2023.

\bibitem[Basmov et~al.(2023)Basmov, Goldberg, and Tsarfaty]{basmov2023chatgpt}
Victoria Basmov, Yoav Goldberg, and Reut Tsarfaty.
\newblock Chatgpt and simple linguistic inferences: Blind spots and blinds, 2023.

\bibitem[Beeching et~al.(2023)Beeching, Han, Lambert, Rajani, Sanseviero, Tunstall, and Wolf]{open-llm-leaderboard}
Edward Beeching, Sheon Han, Nathan Lambert, Nazneen Rajani, Omar Sanseviero, Lewis Tunstall, and Thomas Wolf.
\newblock Open llm leaderboard.
\newblock \url{https://huggingface.co/spaces/HuggingFaceH4/open_llm_leaderboard}, 2023.

\bibitem[Beguš et~al.(2023)Beguš, Dąbkowski, and Rhodes]{beguš2023large}
Gašper Beguš, Maksymilian Dąbkowski, and Ryan Rhodes.
\newblock Large linguistic models: Analyzing theoretical linguistic abilities of llms, 2023.

\bibitem[Bisani \& Ney(2008)Bisani and Ney]{bisani2008joint}
Maximilian Bisani and Hermann Ney.
\newblock Joint-sequence models for grapheme-to-phoneme conversion.
\newblock \emph{Speech communication}, 50\penalty0 (5):\penalty0 434--451, 2008.

\bibitem[Bond-Taylor et~al.(2021)Bond-Taylor, Leach, Long, and Willcocks]{bond2021deep}
Sam Bond-Taylor, Adam Leach, Yang Long, and Chris~G Willcocks.
\newblock Deep generative modelling: A comparative review of vaes, gans, normalizing flows, energy-based and autoregressive models.
\newblock \emph{IEEE transactions on pattern analysis and machine intelligence}, 44\penalty0 (11):\penalty0 7327--7347, 2021.

\bibitem[Bruck \& Genesee(1995)Bruck and Genesee]{bruck1995phonological}
Maggie Bruck and Fred Genesee.
\newblock Phonological awareness in young second language learners.
\newblock \emph{Journal of child Language}, 22\penalty0 (2):\penalty0 307--324, 1995.

\bibitem[Caplan(2014)]{caplan2014rhyme}
David Caplan.
\newblock \emph{Rhyme's challenge: hip hop, poetry, and contemporary rhyming culture}.
\newblock Oxford University Press, USA, 2014.

\bibitem[Chakravarthi et~al.(2019)Chakravarthi, Priyadharshini, Stearns, Jayapal, Sridevy, Arcan, Zarrouk, and McCrae]{chakravarthi2019multilingual}
Bharathi~Raja Chakravarthi, Ruba Priyadharshini, Bernardo Stearns, Arun~Kumar Jayapal, S~Sridevy, Mihael Arcan, Manel Zarrouk, and John~P McCrae.
\newblock Multilingual multimodal machine translation for dravidian languages utilizing phonetic transcription.
\newblock In \emph{Proceedings of the 2nd Workshop on Technologies for MT of Low Resource Languages}, pp.\  56--63, 2019.

\bibitem[Ding et~al.(2024)Ding, Liu, Dong, Zhang, Qian, He, Lin, and Wang]{ding2024songcomposer}
Shuangrui Ding, Zihan Liu, Xiaoyi Dong, Pan Zhang, Rui Qian, Conghui He, Dahua Lin, and Jiaqi Wang.
\newblock Songcomposer: A large language model for lyric and melody composition in song generation, 2024.

\bibitem[Elazar et~al.(2024)Elazar, Bhagia, Magnusson, Ravichander, Schwenk, Suhr, Walsh, Groeneveld, Soldaini, Singh, Hajishirzi, Smith, and Dodge]{elazar2024whats}
Yanai Elazar, Akshita Bhagia, Ian Magnusson, Abhilasha Ravichander, Dustin Schwenk, Alane Suhr, Pete Walsh, Dirk Groeneveld, Luca Soldaini, Sameer Singh, Hanna Hajishirzi, Noah~A. Smith, and Jesse Dodge.
\newblock What's in my big data?, 2024.

\bibitem[Galescu \& Allen(2002)Galescu and Allen]{galescu2002pronunciation}
Lucian Galescu and James~F Allen.
\newblock Pronunciation of proper names with a joint n-gram model for bi-directional grapheme-to-phoneme conversion.
\newblock In \emph{Seventh International Conference on Spoken Language Processing}, 2002.

\bibitem[Henderson(1965)]{henderson1965haiku}
Harold~G Henderson.
\newblock Haiku in english., 1965.

\bibitem[Hu \& Levy(2023)Hu and Levy]{hu2023promptbased}
Jennifer Hu and Roger Levy.
\newblock Prompt-based methods may underestimate large language models' linguistic generalizations, 2023.

\bibitem[Jiang et~al.(2023)Jiang, Sablayrolles, Mensch, Bamford, Chaplot, Casas, Bressand, Lengyel, Lample, Saulnier, et~al.]{jiang2023mistral}
Albert~Q Jiang, Alexandre Sablayrolles, Arthur Mensch, Chris Bamford, Devendra~Singh Chaplot, Diego de~las Casas, Florian Bressand, Gianna Lengyel, Guillaume Lample, Lucile Saulnier, et~al.
\newblock Mistral 7b.
\newblock \emph{arXiv preprint arXiv:2310.06825}, 2023.

\bibitem[Jiang et~al.(2024)Jiang, Sablayrolles, Roux, Mensch, Savary, Bamford, Chaplot, de~las Casas, Hanna, Bressand, Lengyel, Bour, Lample, Lavaud, Saulnier, Lachaux, Stock, Subramanian, Yang, Antoniak, Scao, Gervet, Lavril, Wang, Lacroix, and Sayed]{jiang2024mixtral}
Albert~Q. Jiang, Alexandre Sablayrolles, Antoine Roux, Arthur Mensch, Blanche Savary, Chris Bamford, Devendra~Singh Chaplot, Diego de~las Casas, Emma~Bou Hanna, Florian Bressand, Gianna Lengyel, Guillaume Bour, Guillaume Lample, Lélio~Renard Lavaud, Lucile Saulnier, Marie-Anne Lachaux, Pierre Stock, Sandeep Subramanian, Sophia Yang, Szymon Antoniak, Teven~Le Scao, Théophile Gervet, Thibaut Lavril, Thomas Wang, Timothée Lacroix, and William~El Sayed.
\newblock Mixtral of experts, 2024.

\bibitem[Jiao et~al.(2023)Jiao, Wang, tse Huang, Wang, and Tu]{jiao2023chatgpt}
Wenxiang Jiao, Wenxuan Wang, Jen tse Huang, Xing Wang, and Zhaopeng Tu.
\newblock Is chatgpt a good translator? yes with gpt-4 as the engine, 2023.

\bibitem[Koco{\'{n}} et~al.(2023)Koco{\'{n}}, Cichecki, Kaszyca, Kochanek, Szyd{\l}o, Baran, Bielaniewicz, Gruza, Janz, Kanclerz, Koco{\'{n}}, Koptyra, Mieleszczenko-Kowszewicz, Mi{\l}kowski, Oleksy, Piasecki, Radli{\'{n}}ski, Wojtasik, Wo{\'{z}}niak, and Kazienko]{Koco__2023}
Jan Koco{\'{n}}, Igor Cichecki, Oliwier Kaszyca, Mateusz Kochanek, Dominika Szyd{\l}o, Joanna Baran, Julita Bielaniewicz, Marcin Gruza, Arkadiusz Janz, Kamil Kanclerz, Anna Koco{\'{n}}, Bart{\l}omiej Koptyra, Wiktoria Mieleszczenko-Kowszewicz, Piotr Mi{\l}kowski, Marcin Oleksy, Maciej Piasecki, {\L}ukasz Radli{\'{n}}ski, Konrad Wojtasik, Stanis{\l}aw Wo{\'{z}}niak, and Przemys{\l}aw Kazienko.
\newblock {ChatGPT}: Jack of all trades, master of none.
\newblock \emph{Information Fusion}, pp.\  101861, jun 2023.
\newblock \doi{10.1016/j.inffus.2023.101861}.
\newblock URL \url{https://doi.org/10.1016\%2Fj.inffus.2023.101861}.

\bibitem[Kudo \& Richardson(2018)Kudo and Richardson]{kudo2018sentencepiece}
Taku Kudo and John Richardson.
\newblock Sentencepiece: A simple and language independent subword tokenizer and detokenizer for neural text processing.
\newblock \emph{arXiv preprint arXiv:1808.06226}, 2018.

\bibitem[Kwon(2023)]{kwon2023interfaces}
Taeahn Kwon.
\newblock \emph{Interfaces for Personalized Language Learning with Generative Language Models}.
\newblock PhD thesis, Columbia University, 2023.

\bibitem[Liang et~al.(2022)Liang, Bommasani, Lee, Tsipras, Soylu, Yasunaga, Zhang, Narayanan, Wu, Kumar, Newman, Yuan, Yan, Zhang, Cosgrove, Manning, Ré, Acosta-Navas, Hudson, Zelikman, Durmus, Ladhak, Rong, Ren, Yao, Wang, Santhanam, Orr, Zheng, Yuksekgonul, Suzgun, Kim, Guha, Chatterji, Khattab, Henderson, Huang, Chi, Xie, Santurkar, Ganguli, Hashimoto, Icard, Zhang, Chaudhary, Wang, Li, Mai, Zhang, and Koreeda]{liang2022holistic}
Percy Liang, Rishi Bommasani, Tony Lee, Dimitris Tsipras, Dilara Soylu, Michihiro Yasunaga, Yian Zhang, Deepak Narayanan, Yuhuai Wu, Ananya Kumar, Benjamin Newman, Binhang Yuan, Bobby Yan, Ce~Zhang, Christian Cosgrove, Christopher~D. Manning, Christopher Ré, Diana Acosta-Navas, Drew~A. Hudson, Eric Zelikman, Esin Durmus, Faisal Ladhak, Frieda Rong, Hongyu Ren, Huaxiu Yao, Jue Wang, Keshav Santhanam, Laurel Orr, Lucia Zheng, Mert Yuksekgonul, Mirac Suzgun, Nathan Kim, Neel Guha, Niladri Chatterji, Omar Khattab, Peter Henderson, Qian Huang, Ryan Chi, Sang~Michael Xie, Shibani Santurkar, Surya Ganguli, Tatsunori Hashimoto, Thomas Icard, Tianyi Zhang, Vishrav Chaudhary, William Wang, Xuechen Li, Yifan Mai, Yuhui Zhang, and Yuta Koreeda.
\newblock Holistic evaluation of language models, 2022.

\bibitem[Liu et~al.(2019)Liu, Ma, Huang, Xiong, and He]{liu-etal-2019-robust}
Hairong Liu, Mingbo Ma, Liang Huang, Hao Xiong, and Zhongjun He.
\newblock Robust neural machine translation with joint textual and phonetic embedding.
\newblock In \emph{Proceedings of the 57th Annual Meeting of the Association for Computational Linguistics}, pp.\  3044--3049, Florence, Italy, July 2019. Association for Computational Linguistics.
\newblock \doi{10.18653/v1/P19-1291}.
\newblock URL \url{https://aclanthology.org/P19-1291}.

\bibitem[Maguire(2006)]{maguire2006american}
James Maguire.
\newblock \emph{American bee: The national spelling bee and the culture of word nerds}.
\newblock Rodale Books, 2006.

\bibitem[Masumura et~al.(2020)Masumura, Makishima, Ihori, Takashima, Tanaka, and Orihashi]{masumura2020phoneme}
Ryo Masumura, Naoki Makishima, Mana Ihori, Akihiko Takashima, Tomohiro Tanaka, and Shota Orihashi.
\newblock Phoneme-to-grapheme conversion based large-scale pre-training for end-to-end automatic speech recognition.
\newblock In \emph{INTERSPEECH}, pp.\  2822--2826, 2020.

\bibitem[Mikk(2008)]{mikk2008sentence}
Jaan Mikk.
\newblock Sentence length for revealing the cognitive load reversal effect in text comprehension.
\newblock \emph{Educational Studies}, 34\penalty0 (2):\penalty0 119--127, 2008.

\bibitem[OpenAI(2022)]{chatgpt2022}
OpenAI.
\newblock Introducing chatgpt.
\newblock 2022.
\newblock URL \url{https://openai.com/blog/chatgpt}.

\bibitem[OpenAI(2023)]{openai2023gpt4}
OpenAI.
\newblock Gpt-4 technical report, 2023.

\bibitem[Ormazabal et~al.(2022)Ormazabal, Artetxe, Agirrezabal, Soroa, and Agirre]{ormazabal2022poelm}
Aitor Ormazabal, Mikel Artetxe, Manex Agirrezabal, Aitor Soroa, and Eneko Agirre.
\newblock Poelm: A meter-and rhyme-controllable language model for unsupervised poetry generation.
\newblock \emph{arXiv preprint arXiv:2205.12206}, 2022.

\bibitem[Park \& Kim(2019)Park and Kim]{g2pE2019}
Kyubyong Park and Jongseok Kim.
\newblock g2pe.
\newblock \url{https://github.com/Kyubyong/g2p}, 2019.

\bibitem[Peng et~al.(2023)Peng, Nuchged, and Gao]{peng2023spoken}
Linkai Peng, Baorian Nuchged, and Yingming Gao.
\newblock Spoken language intelligence of large language models for language learning, 2023.

\bibitem[Qin et~al.(2023)Qin, Zhang, Zhang, Chen, Yasunaga, and Yang]{qin2023chatgpt}
Chengwei Qin, Aston Zhang, Zhuosheng Zhang, Jiaao Chen, Michihiro Yasunaga, and Diyi Yang.
\newblock Is chatgpt a general-purpose natural language processing task solver?, 2023.

\bibitem[Raffel et~al.(2023)Raffel, Shazeer, Roberts, Lee, Narang, Matena, Zhou, Li, and Liu]{raffel2023exploring}
Colin Raffel, Noam Shazeer, Adam Roberts, Katherine Lee, Sharan Narang, Michael Matena, Yanqi Zhou, Wei Li, and Peter~J. Liu.
\newblock Exploring the limits of transfer learning with a unified text-to-text transformer, 2023.

\bibitem[Razeghi et~al.(2022)Razeghi, Logan~IV, Gardner, and Singh]{razeghi2022impact}
Yasaman Razeghi, Robert~L Logan~IV, Matt Gardner, and Sameer Singh.
\newblock Impact of pretraining term frequencies on few-shot reasoning.
\newblock \emph{arXiv preprint arXiv:2202.07206}, 2022.

\bibitem[Ruthotto \& Haber(2021)Ruthotto and Haber]{ruthotto2021introduction}
Lars Ruthotto and Eldad Haber.
\newblock An introduction to deep generative modeling.
\newblock \emph{GAMM-Mitteilungen}, 44\penalty0 (2):\penalty0 e202100008, 2021.

\bibitem[Sawicki et~al.(2023)Sawicki, Grzes, Goes, Brown, Peeperkorn, and Khatun]{sawicki2023bits}
Piotr Sawicki, Marek Grzes, Fabricio Goes, Dan Brown, Max Peeperkorn, and Aisha Khatun.
\newblock Bits of grass: Does gpt already know how to write like whitman?
\newblock \emph{arXiv preprint arXiv:2305.11064}, 2023.

\bibitem[Shibata et~al.(1999)Shibata, Kida, Fukamachi, Takeda, Shinohara, Shinohara, and Arikawa]{shibata1999byte}
Yusuxke Shibata, Takuya Kida, Shuichi Fukamachi, Masayuki Takeda, Ayumi Shinohara, Takeshi Shinohara, and Setsuo Arikawa.
\newblock Byte pair encoding: A text compression scheme that accelerates pattern matching.
\newblock 1999.

\bibitem[Sun et~al.(2023)Sun, Tian, Zhou, Xu, Hu, Gupta, Wieting, Peng, and Ma]{sun2023evaluating}
Jiao Sun, Yufei Tian, Wangchunshu Zhou, Nan Xu, Qian Hu, Rahul Gupta, John~Frederick Wieting, Nanyun Peng, and Xuezhe Ma.
\newblock Evaluating large language models on controlled generation tasks, 2023.

\bibitem[Suzuki(1985)]{suzuki1985role}
Seiichi Suzuki.
\newblock The role of syllable structure in old english poetry.
\newblock \emph{Lingua}, 67\penalty0 (2-3):\penalty0 97--119, 1985.

\bibitem[Tian \& Peng(2022)Tian and Peng]{tian2022zeroshot}
Yufei Tian and Nanyun Peng.
\newblock Zero-shot sonnet generation with discourse-level planning and aesthetics features, 2022.

\bibitem[Tian et~al.(2023)Tian, Narayan-Chen, Oraby, Cervone, Sigurdsson, Tao, Zhao, Chen, Chung, Huang, and Peng]{tian2023unsupervised}
Yufei Tian, Anjali Narayan-Chen, Shereen Oraby, Alessandra Cervone, Gunnar Sigurdsson, Chenyang Tao, Wenbo Zhao, Yiwen Chen, Tagyoung Chung, Jing Huang, and Nanyun Peng.
\newblock Unsupervised melody-to-lyric generation, 2023.

\bibitem[Toshniwal \& Livescu(2016)Toshniwal and Livescu]{toshniwal2016jointly}
Shubham Toshniwal and Karen Livescu.
\newblock Jointly learning to align and convert graphemes to phonemes with neural attention models.
\newblock In \emph{2016 IEEE Spoken Language Technology Workshop (SLT)}, pp.\  76--82. IEEE, 2016.

\bibitem[Ukrainetz et~al.(2011)Ukrainetz, Nuspl, Wilkerson, and Beddes]{ukrainetz2011effects}
Teresa~A Ukrainetz, Janae~J Nuspl, Kimberly Wilkerson, and Sarah~Rose Beddes.
\newblock The effects of syllable instruction on phonemic awareness in preschoolers.
\newblock \emph{Early childhood research quarterly}, 26\penalty0 (1):\penalty0 50--60, 2011.

\bibitem[Valmeekam et~al.(2023)Valmeekam, Olmo, Sreedharan, and Kambhampati]{valmeekam2023large}
Karthik Valmeekam, Alberto Olmo, Sarath Sreedharan, and Subbarao Kambhampati.
\newblock Large language models still can't plan (a benchmark for llms on planning and reasoning about change), 2023.

\bibitem[Vice(2023)]{vice_poem}
Vice.
\newblock Chatgpt is pretty bad at poetry, according to poets.
\newblock \url{https://www.vice.com/en/article/7kx9d9/chatgpt-is-pretty-bad-at-poetry-according-to-poets}, 2023.

\bibitem[Yolchuyeva et~al.(2019)Yolchuyeva, Németh, and Gyires-Tóth]{Yolchuyeva_2019}
Sevinj Yolchuyeva, Géza Németh, and Bálint Gyires-Tóth.
\newblock Transformer based grapheme-to-phoneme conversion.
\newblock In \emph{Interspeech 2019}, interspeech2019. ISCA, September 2019.
\newblock \doi{10.21437/interspeech.2019-1954}.
\newblock URL \url{http://dx.doi.org/10.21437/Interspeech.2019-1954}.

\bibitem[Yu et~al.(2024)Yu, Zang, Wang, Zhuang, and Gu]{yu2024token}
Chengyue Yu, Lei Zang, Jiaotuan Wang, Chenyi Zhuang, and Jinjie Gu.
\newblock Token-free llms can generate chinese classical poetry with more accurate format.
\newblock \emph{arXiv preprint arXiv:2401.03512}, 2024.

\bibitem[Zheng et~al.(2023)Zheng, Chiang, Sheng, Zhuang, Wu, Zhuang, Lin, Li, Li, Xing, Zhang, Gonzalez, and Stoica]{zheng2023judging}
Lianmin Zheng, Wei-Lin Chiang, Ying Sheng, Siyuan Zhuang, Zhanghao Wu, Yonghao Zhuang, Zi~Lin, Zhuohan Li, Dacheng Li, Eric.~P Xing, Hao Zhang, Joseph~E. Gonzalez, and Ion Stoica.
\newblock Judging llm-as-a-judge with mt-bench and chatbot arena, 2023.

\end{thebibliography}
